# Using the SLEUTH urban growth model to simulate the impacts of future policy scenarios on urban land use in the Tehran metropolitan area in Iran

Shaghayegh Kargozar Nahavandy[a,*], Lalit Kumar[b], and Pedram Ghamisi[c]

[a] Department of Surveying and Geomatics, Faculty of Engineering, Tehran University, Tehran, Iran

[b] Ecosystem Management, School of Environmental and Rural Science, University of New England, Armidale NSW 2351, Australia

[c] German Aerospace Center (DLR), Remote Sensing Technology Institute (IMF) and Technical University of Munich (TUM), Signal Processing in Earth Observation, Munich, Germany

(corresponding author, e-mail: p.ghamisi@gmail.com).

## ABSTRACT

The SLEUTH model, based on the Cellular Automata (CA), can be applied to city development simulation in metropolitan areas. In this study the SLEUTH model was used to model the urban expansion and predict the future possible behavior of the urban growth in Tehran. The fundamental data were five Landsat TM and ETM images of 1988, 1992, 1998, 2001 and 2010. Three scenarios were designed to simulate the spatial pattern. The first scenario assumed historical urbanization mode would persist and the only limitations for development were height and slope. The second one was a compact scenario which makes the growth mostly internal and limited the expansion of suburban areas. The last scenario proposed a polycentric urban structure which let the little patches grow without any limitation and would not consider the areas beyond the specific buffer zone from the larger patches for development. Results showed that the urban growth rate was greater in the first scenario in comparison with the other two scenarios. Also it was shown that the third scenario was more suitable for Tehran since it could avoid undesirable effects such as congestion and pollution and was more in accordance with the conditions of Tehran city.

* Corresponding author. Tel.: +98 912 3572913
  E-mail address: shaghayegh.kargozar@yahoo.com

**Keywords:** Urban growth, Cellular Automata (CA), SLEUTH

# 1. INTRODUCTION

In recent years, with increasing population around the world and all the consequences of this increment, urbanization simulation processes are rapidly gaining popularity among urban planners and geographers. Urbanization is the conversion from natural to artificial land cover characterized by human settlements and workplaces [1]. Dynamic spatial urban models that provide an improved ability to assess future growth and create planning scenarios, allows the exploration of impacts of decisions that follow different urban planning and management policies [2-4].

In the last few decades, models based on Cellular Automata (CA) have been claimed to be useful tools for modeling the urban phenomena [5,6]. CA are dynamic discrete space and time systems. In classic CA, the space is defined as a regular lattice of cells and the state of each cell updates synchronously in discrete time steps in respect to its local relationships. CA have many advantages for modeling urban phenomena, including their decentralized approach, the link they provide to the complexity theory, the relative ease with which model results can be visualized, their flexibility, their dynamic approach and also their affinities with geographical information systems and remotely sensed data [7]. The many applications of CA have shown that CA offers a flexible and advanced spatial modeling environment that has not been available before [8-13].

In this research, we used a CA based model called SLEUTH (Slope, Land cover, Exclusion, Urban, Transportation, and Hillshade) to simulate the urban expansion of Tehran city and investigate the consequences of different growth conditions. This model was developed by Keith Clarke and co-authors in the mid-1990s [1,14,15]. Urban development view of SLEUTH model does not only assume unitary cells, but also diffusion of more complex urban entities as a whole [16]. SLEUTH can be used under different conditions and has the ability to simulate different urban plans and socioeconomic policies [17-20]. Thus, in metropolitan areas like Tehran city, this model can be useful for investigating the consequences of decisions made by urban planners and helping them design suitable development plans in order to control the irregular and unexpected growth. The aims of this paper are: (1) to simulate urban expansion in Tehran city using SLEUTH; (2) to investigate the consequences of different growth conditions in the manner of development using different growth scenarios; and (3) to propose the most appropriate method for controlling the urban growth and managing the city in accordance with the urban policy.

## 2. MATERIALS AND METHODS

### 2.1 Characteristics of the study area

Tehran is currently one of the largest and most populated cities of the world. It has gone through a lot of upheavals in history. Tehran, which initially was a small village, developed gradually and became a metropolis with a population of 4.2 million in 1974 and 12.1 million by 2010. It was selected as the Capital of Iran in 1785 by the first king of Ghajar [21,22], and since then has been the political, cultural, economic and commercial nucleus of the country.

In 1867, the population of Tehran was estimated to be 147256. Table 1 shows the variation of growth rate during the last five centuries. The onset of increase in population is assigned to early Pahlavi era. The population increased by about ten times greater during the next 40 years and reached 2 million by 1961. The increase was more gradual over the next four decades and eventually reached 7.7 million in 2006. From Table 1, it is obvious that the population growth rate decreased from 1966 to 2006, although five million people were added to the population. This increase had socioeconomic and bioenvironmental consequences. The increase in population in cities around Tehran was much faster compared to central Tehran. In the other words, the growth of the population in Tehran has decreased in recent decades; however the population of the surrounding cities has increased.

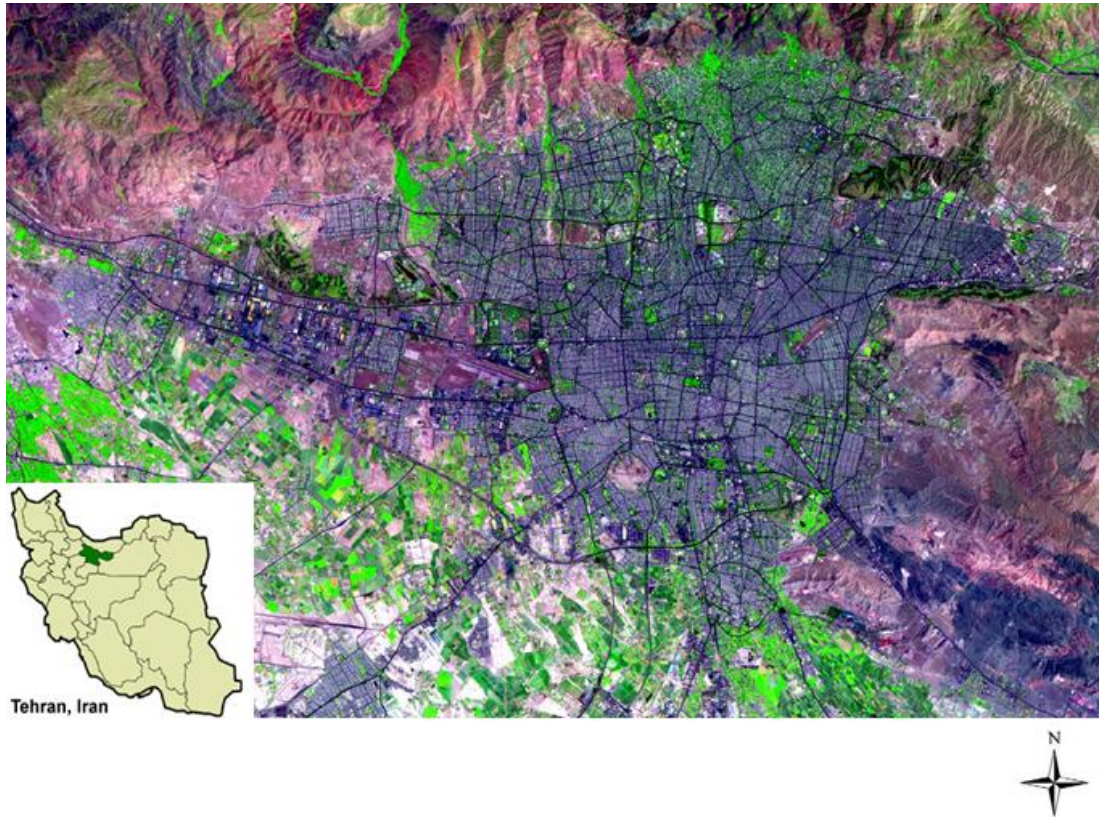

Figure 1 Location of Tehran in Iran. The urban areas are shown in blue color.

Table 1 The population and growth rate information of Tehran in the last five centuries [23]

| Year | Population | Growth rate (%) | Government |
|---|---|---|---|
| 1554 | 1000 | | Shah Tahmasb-e Safavi |
| 1626 | 3000 | 1.4 | Shah Abbas Safavi |
| 1797 | 15000 | 5.2 | Agha Mohammad Khan Ghajar |
| 1807 | 50000 | 12.03 | Fathali Shah-e Ghajar |
| 1812 | 60000 | 3.6 | Fathali Shah-e Ghajar |
| 1834 | 80000 | 2.8 | Fathali Shah-e Ghajar |
| 1867 | 147256 | 2.9 | Nasereddin Shah |
| 1930 | 250000 | 2.4 | First Pahlavi |
| 1940 | 540087 | 6.6 | First Pahlavi |
| 1956 | 1560934 | 5.5 | Second Pahlavi |
| 1966 | 2719730 | 5.1 | Second Pahlavi |
| 1976 | 4530223 | 2.9 | Second Pahlavi |
| 1986 | 6058207 | 1.3 | Islamic republic of Iran |
| 1991 | 6497238 | 0.78 | Islamic republic of Iran |
| 1996 | 6758845 | 1.3 | Islamic republic of Iran |
| 2006 | 7711230 | 1.4 | Islamic republic of Iran |

**2.2 Model description**

SLEUTH is a modified Cellular Automata (CA) model [1]. This model is composed of two main submodels; Urban Growth Model (UGM) and Deltatron Land use Model (DLM). As the standard CA, SLEUTH needs to be initialized to be ready for applying the transition rules. The input requirements of SLEUTH are the layers describing slope, land cover, excluded areas, existing urbanized areas and transportation networks. Changes in urban form are implemented in four substeps in the model; these substeps are the growth rules describing the different forms of spread and diffusion in space. The first substep describes spontaneous urbanization. In this step, each cell has the potential to be transformed into urban state and is determined by the slope at that cell. In the second substep, new diffusion centers are generated. When each spontaneously urbanized cell (C) has two or more non-urbanized adjacent cells within a 3×3 Moore neighborhood [16], there is a fixed probability that C becomes a new spreading center and urbanize two of its neighbors. At substep 3, each urbanized cell can grow on their edges. This means that each edge cell that has three or more urbanized neighbors within its 3×3 Moore neighborhood has a fixed probability to become urban. The fourth substep describes further development inspired by an urban cell close to a road. After selecting a cell close to the road as a new spreading center, it is transported along the road in a randomly selected direction for a walk of fixed length and is anchored at the destination.

Furthermore, there are five parameters controlling the behavior of the system. These parameters affect the growth rules that determine the forms of urbanization. Each coefficient may be an integer between 0 and 100. Comparing the simulated land cover change to a study area's historical data and calculating linear regression, goodness-of-fit scores ($r^2$) calibrate these values. These parameters are Dispersion coefficient, Breed coefficient, Spread coefficient, Slope-Resistance coefficient and Road-Gravity coefficient. The Dispersion coefficient affects the spontaneous growth by controlling the number of times a pixel will be randomly selected for possible urbanization. This coefficient also controls the number of pixels that make up a random walk along the transportation network as part of road-influenced growth. The Breed coefficient affects new spreading center growth and road-influenced growth. This factor determines the probability of a pixel being urbanized by spontaneous growth, becoming a new spreading center and indicating the number of times a road trip will be taken during road-influenced growth. The Spread coefficient only influences the edge growth by determining the probability that any pixel that has become a spreading center before will produce an additional urban pixel in its neighborhood. The Slope-Resistance coefficient checks the slope of each cell and affects all the growth rules by making sure that the location of cells is suitable for urbanization. Finally the Road-Gravity coefficient determines the maximum search distance from a pixel selected for a road trip to a road pixel. After each time step, each one of these parameters will be modified by the self-modification constants. In fact, this self-modification behavior of the model alters the coefficient values to simulate the growth.

The SLEUTH model would not be capable of representing the growth of real urban areas if the model parameters remained fixed during the modeling period. Hence the main purpose of the model is to find the best values of the parameters to make the model capable of simulating real-world cities. In this process, when the growth rate exceeds a given threshold, the probability of spontaneous urbanization and all diffusion rates are multiplied by a factor greater than one. To prevent explosive growth, acceleration is controlled, and the multiplier is decreased linearly with the aging of a cluster. In a similar way, when the cluster growth rate decreases below the other threshold, the growth and diffusion rates are slowed down even more, multiplied by a factor less than one. Once aging, to prevent collapse, these parameters increase linearly with the age of a cluster [1].

**2.3 Preparing the input layers**

Implementation of the model occurs in two main phases: (1) calibration, and (2) prediction. For calibration, the model requires historic urban extent for at least four time periods, a historic transportation network for at least two time periods, slope, hillshade and an excluded layer. All of the input layers were derived from satellite images. Urban layers and transportation layers were generated by classifying the Landsat Thematic mapper (TM) and Enhanced Thematic mapper (ETM) imageries with resolution of 30 meters. Five time steps for these layers were prepared for calibration of the model; 1988, 1992, 1998, 2001 and 2010.

For the classification stage, Support Vector Machine (SVM) method was taken into account. SVM separates training samples which belong to different classes by tracing maximum margin hyperplanes in the space where the samples are mapped [27]. SVMs were originally introduced to solve linear classification problems. However, they can be generalized to non-linear decision functions by considering the so-called kernel trick [28]. A kernel-based SVM is exploited to project the pixel vectors into a higher dimensional space and estimate maximum margin hyperplanes in this new space for improving linear separability of data [29]. The sensitivity to the choice of the kernel and regularization parameters are considered as the most important shortcomings of SVM. In order to address the first shortcoming, Radial Basis Function (RBF) kernel has been widely used in remote sensing. The latter can be addressed by considering cross-validation techniques using training data in order to tune hyperplane parameters [29].

Classification accuracy assessments were undertaken for each image to determine how well the images were classified. The classified images were then converted to binary images, with 1 representing urban/road areas and 0 representing the other areas. In order to generate more accurate road layers, some ancillary operations were implemented in road classes in the classified images. As an example, due to the fact that some main roads that connect the main body of the city to the suburbs, especially in earlier years, are made of the same material as arid regions, they need revision to gain more precise results. Hence accurate paper road maps for the selected years were digitized and rescaled to be used as auxiliary data for modification of the road layers.

The slope and hillshade layers were derived from a Digital Elevation Model (DEM) of Tehran, having a spatial resolution of 30 m. In the slope layer, the cell values were calculated in percent slope, with 0 representing flat surfaces and 100 representing 90 degree slopes. In order to give spatial context to the urban extent data, a background image was incorporated into the image output. The Hillshaded layer was used for this purpose. The excluded layer consisted of areas with prohibition of urbanization, such as water, green spaces and sleep areas. The input datasets are given in Table 2.

**2.4 Calibration and prediction**

The main purpose of calibration is to obtain the best set of parameters to make the model capable of simulating a city with the best similarity with the reality. A secure method to reach that goal is testing all possible permutations of the parameters to find the best set of coefficients. Obviously, this process requires fast CPU and takes a long time to explore parameter values. To overcome this problem, the methodology of brute force calibration was utilized to derive parameter values. In this method, instead of executing every combination of possible coefficient sets, each coefficient range is examined in increments. The process is implemented in three phases; coarse calibration, fine calibration and final calibration. In each phase, the coefficient range, increment size and resolution of the input layers were changed. In other words, as the parameter range was narrowed down in each phase, the increment size became smaller and the resolution of the input images became better and closer to the full image resolution as in final calibration phase, the increment size was the lowest and the full resolution images were used as input layers. In this way, the number of solution sets was reduced, but still the full range of solutions was searched to find the best fit parameters.

Table 2 List of input datasets

| Input Data | Source | year |
|---|---|---|
| Urban Layers | Classified from Landsat image ( TM 30m) | 1988,1992,1998,2001,2010 |
| Transportation Networks | Classified from Landsat image ( TM 30m) | 1988,1992,1998,2001,2010 |
| Slope Layer | Aster DEM | 1996 |
| Hillshade Layer | Aster DEM | 1996 |
| Excluded Layer | Generated by the program | 2010 |

Due to the high level of randomness present in each growth cycle, growth simulations were generated in Monte Carlo fashion to bring a greater amount of stability to modeled results. Monte Carlo averaging reduces dependence upon the initial conditions [25]. Each phase of calibration was implemented in 100 Monte Carlo iterations and the final results of each coefficient set would be the average of them during the Monte Carlo iterations.

In order to find the best fit model, 11 different metrics were calculated for each run of the model (Table 3) so that by comparing the historical metrics with them, linear regression values could be calculated and the best fit values could be used for selecting the best coefficient set during the calibration process; though with the increment of the number of calibration metrics, it would be harder to find the factors that influence the calibration scores more than the others. Likewise, the product of these metrics would have the same problem. In this research, a shape factor called Leesallee was used for choosing the best parameter set. Leesallee is a measurement of spatial fit between the urban growth that is modeled and the known urban extent for the control years; 1 representing a perfect match and 0 representing a spatial disconnect [26].

After the final phase of the calibration, the best fit parameter set that is calculated in calibration phases would be used in order to produce the best parameters for the final date. This parameter set was used to initialize the forecasting phase by running the model for 100 Monte Carlo iterations and averaging the generated parameters. The best fit parameters for the prediction phase are shown in Table 4.

The prediction phase of the SLEUTH model simulates the future aspects of the city, using historical growth trends from the past. The best fit parameters of the last historical data that is obtained in the final calibration phase, were used to initialize the forecasting phase, and then the growth rules were applied to the data for the specific number of years. In fact, the parameters of the most recent year were used to initialize the forecast run and then the prediction was implemented using best fit calibration parameters to accomplish the process.

In order to investigate the consequences of different urban policies on the impact and mode of urbanization, various scenarios can be used for simulating different conditions. Comparing the results of these scenarios produces practical information for urban planners to avoid unexpected growth and keep the development under control. There are two methods to simulate different situations in the model. The first one is by changing the parameter values that affect the trend of urban growth [20], and the second one is by changing the excluded layer to prevent the nonresidential areas that must be preserved from urbanization [24]. In this research, the second method has been used and Tehran province was simulated using 3 scenarios. The first scenario simulated the urban growth with only the limitation of slope and height, and without any other exterior exclusion, and the historical growth trend was the only important factor to implement the prediction. The second scenario was a compact scenario. In this scenario the little patches in suburb areas were not allowed to grow. The goal of this scenario was to prevent the rural areas from urbanization, because by developing the suburb villages around the main body of the city, it would be probable for them to become a part of the city after a few decades. This development would be much faster than the main body growth, due to the incorporation of the rural areas. The third scenario incorporated a polycentric urban structure which allowed the growth to spread all over the province. In this scenario, the boundaries were excluded from urbanization and the most development potential was given to the suburb areas and villages to take the centralization away from the main city and, as a result, reduce the population density and pollution of the developing city.

Table 3 Eleven metrics for evaluation of calibration in SLEUTH model

| Index | Description |
|---|---|
| Compare | compares the amount of modeled urban area to known urban area for the stop date year |
| Pop | least squares regression score for modeled urban area compared to actual urban area for the control years |
| Edges | least squares regression score for the modeled amount of urban perimeter, or edge, compared to actual urban perimeter for the control years |
| Clusters | least squares regression score for modeled number of urban clusters compared to known number of urban clusters for the control years |
| Cluster_size | least squares regression score for modeled average urban cluster size compared to known average urban cluster size for the control years |
| Leesallee | a shape index, a measurement of spatial fit between the model's growth and the known urban extent for the control years, 1 being a perfect match and 0 representing a spatial disconnect |
| Slope | least squares regression of average slope for modeled urbanized cells compared to average slope of known urban |
|        | cells for the control years                                                                                                                                                     |
| %Urban | least squares regression of percent of available pixels urbanized compared to the urbanized pixels for the control years                                                         |
| Xmean  | least squares regression of average longitude (calculated using columnvalues) for modeled urbanized locations compared to average longitude of known urban locations for the control years |
| Ymean  | least squares regression of average latitude (calculated using row values) compared to average latitude of known urban locations for the control years                           |
| Rad    | $\sqrt{std_x^2 + std_y^2}$                                                                                                                                                      |

Table 4 The best fit prediction parameters

| Diffusion | Spread | Breed | Slope Resistance | Road Gravity |
|-----------|--------|-------|------------------|--------------|
| 1         | 36     | 3     | 34               | 49           |

100 Monte Carlo iterations were used for the prediction in each one of the three scenarios to make the results more stable and, in each scenario, the prediction phase was implemented and the urban area for the next 20 years was simulated and annual urban growth maps were produced. These three scenarios were used in this research to show the usefulness of the model in "what-if" experiments, and provide an overall situation for managers to compare the results of different development policies.

## 3. RESULTS

As mentioned in section 2.3, the urban and road layers were extracted from the classified images of years 1988, 1992, 1998, 2001 and 2010. The classified images and the extracted urban and road layers are shown in Figures 2-4. Classification accuracy assessments were undertaken for each image using kappa coefficient and overall accuracy. The results of the classification accuracy are shown in Table 5.

Table 5 Classification accuracy of the Landsat images.

| Year | 1988 | 1992 | 1998 | 2001 | 2010 |
|---|---|---|---|---|---|
| Overall Accuracy | 92.55% | 96.59% | 96.09% | 94.64% | 96.94% |
| Kappa Coefficient | 0.8752 | 0.9414 | 0.9333 | 0.9105 | 0.951 |

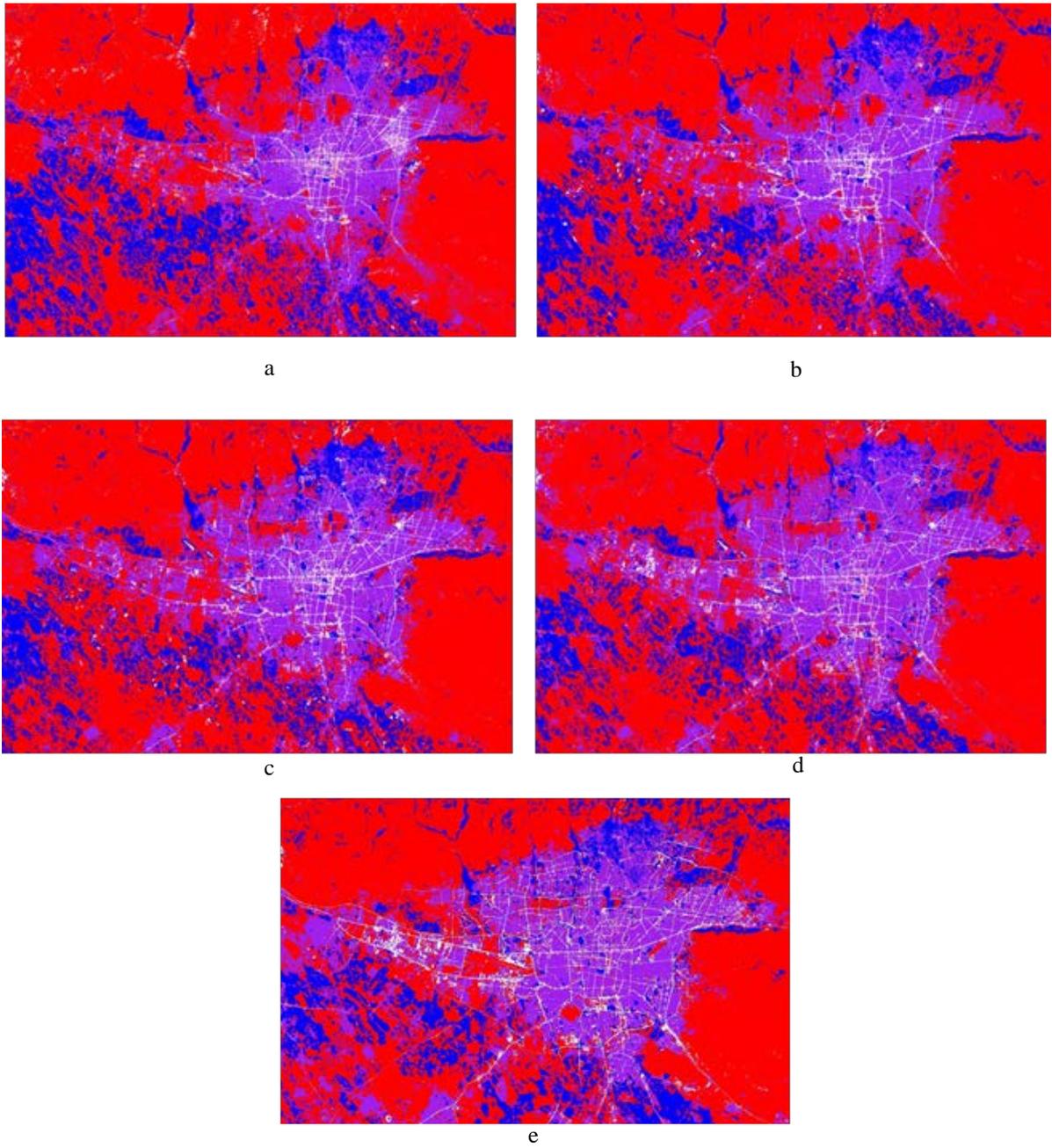

Figure 2 Classified images; (a) 1988, (b) 1992, (c) 1998, (d) 2001, (e) 2010

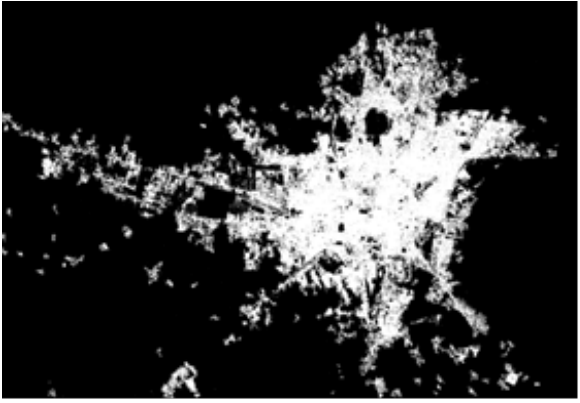
a
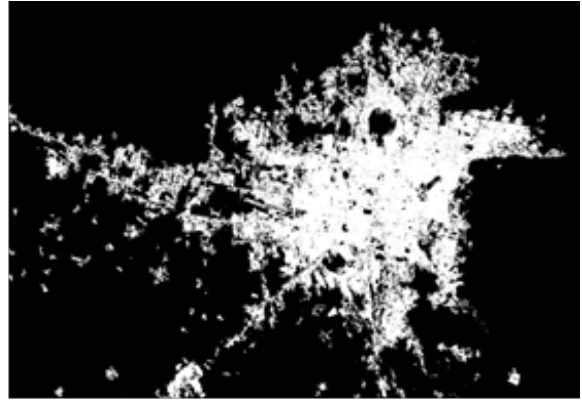
b
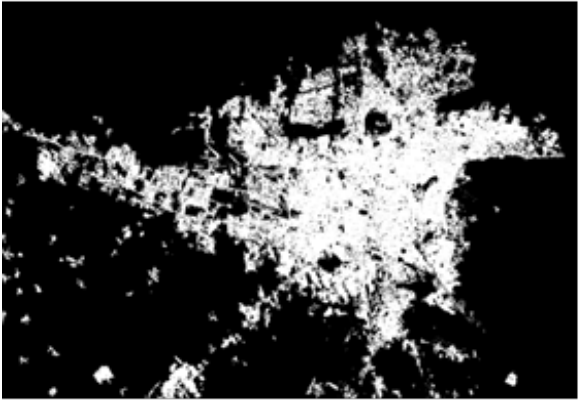
c
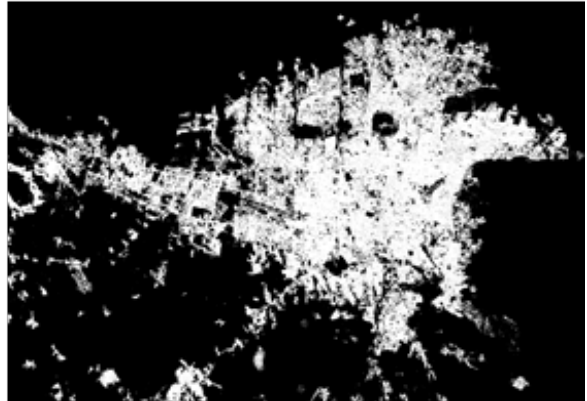
d
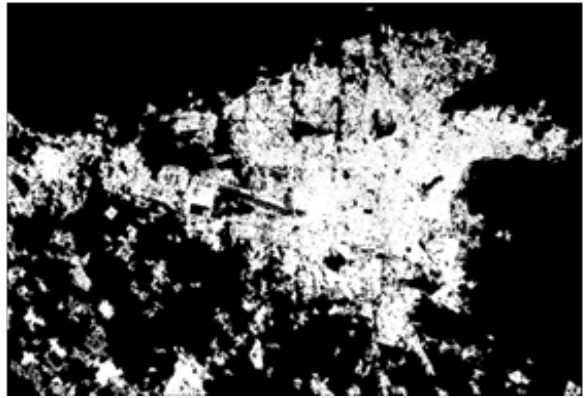
e

Figure 3 Urban layers; (a) 1988, (b) 1992, (c) 1998, (d) 2001, (e) 2010

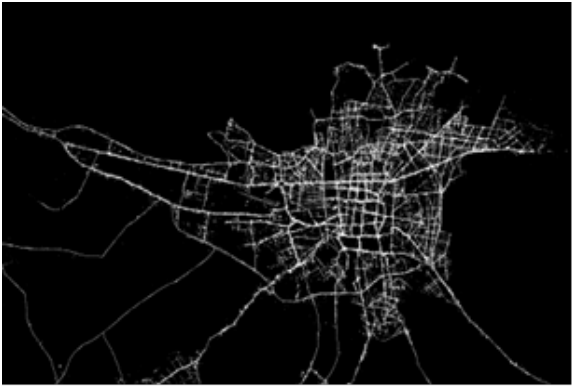
a
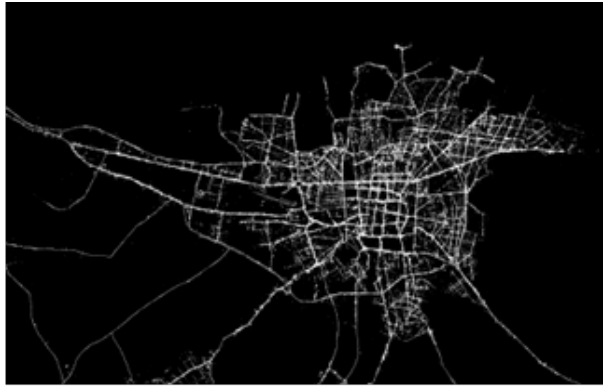
b
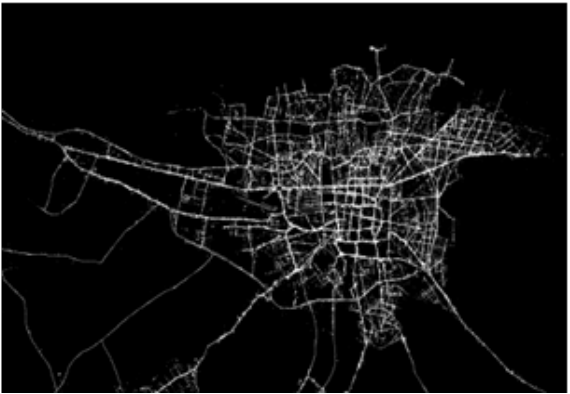
c
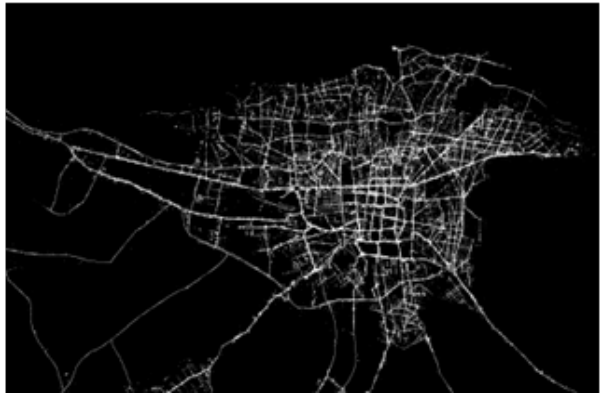
d
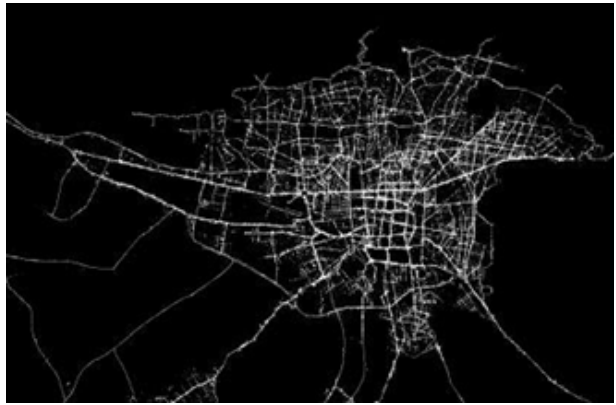
e

Figure 4 Road layers; (a) 1988, (b) 1992, (c) 1998, (d) 2001, (e) 2010

After preparing the input layers and executing the calibration process and prediction phase, the output images were acquired as urban growth probabilistic maps for each year from 2011 to 2030. In the first scenario, as can be seen in Figure 5a, since the only limitations for urbanization were height and slope, development can be seen in the interior parts of the city, boundaries and the suburb areas; though in the north and northeast parts, no growth can be seen. The reason for this is the existence of mountainous regions in the northern parts of the province with their highlands and steep slopes which were excluded from urbanization in the excluded layer. Furthermore, since there were more developable lands in the west and northwest of the city rather than east of the city, and also because of the considerable growth trend in those areas according to historical data, the model was executed in a way to encourage the urbanization in those areas, hence urban development seems more significant in the mentioned areas rather than the eastern part. In the second scenario, in the rural and suburb areas, no development can be observed and only the main central part of the city was urbanized (Figure 5b). In the last scenario, the possibility of growth was taken from boundaries and, as a result, the urbanization only occurred in the interior parts of the city and mostly occurred in the rural areas (Figure 5c)

Some of the most important statistical measures for the three scenarios in the prediction period are shown in Table 6 in order to compare the characteristics of urban areas in the three scenarios. As can be seen from this table, Spread coefficients are greater in the first scenario than the others, but Diffusion coefficient values that represent the urban dispersion, and Breed and Slope-resistance coefficients, did not change considerably during the three scenarios.

Table 6 Statistical measures for three scenarios in years 2010, 2020 and 2030

| Statistical measures | 2010 | | | 2020 | | | 2030 | | |
|---|---|---|---|---|---|---|---|---|---|
| | S1 | S2 | S3 | S1 | S2 | S3 | S1 | S2 | S3 |
| Diffuse | 1.08 | 1.08 | 1.08 | 1.12 | 1.08 | 1.08 | 1.12 | 1.08 | 1.08 |
| Spread | 38.98 | 48.73 | 50.89 | 40.24 | 48.73 | 50.89 | 40.24 | 48.73 | 50.89 |
| Breed | 3.25 | 3.25 | 3.25 | 3.35 | 3.25 | 3.25 | 3.35 | 3.25 | 3.25 |
| slp_res | 1 | 1 | 1 | 1 | 1 | 1 | 1 | 1 | 1 |
| rd_grav | 52.96 | 46.48 | 49.32 | 54.7 | 46.48 | 49.32 | 54.7 | 46.48 | 49.32 |
| Sng | 4.14 | 2.6 | 1.12 | 4.18 | 2.56 | 0.78 | 3.77 | 2.25 | 0.71 |
| Og | 11292 | 8675 | 8550 | 6464 | 4271 | 3966 | 5022 | 3515 | 2719 |
| Rt | 2.55 | 0.74 | 0.82 | 2.84 | 1.08 | 1.48 | 1.42 | 1.28 | 1.02 |
| Area | 602098 | 595426 | 603585 | 684400 | 650103 | 658068 | 739900 | 687820 | 689896 |
| Xmean | 910.37 | 907.4 | 919.24 | 893.17 | 887.07 | 909.45 | 879.1 | 870.07 | 902.75 |
| Ymean | 526.49 | 529.31 | 510.87 | 536.64 | 543.25 | 510.42 | 546.13 | 555.98 | 510.41 |
| %urban | 53.91 | 73.58 | 98.6 | 60.12 | 79.26 | 106.09 | 64.3 | 83.17 | 110.47 |
| grw_rate | 1.88 | 1.37 | 1.42 | 0.95 | 0.66 | 0.6 | 0.68 | 0.51 | 0.39 |
| grw_pix | 11299 | 8178 | 8552 | 6471 | 4275 | 3968 | 5028 | 3519 | 2721 |

The *sng* parameter in Table 6 is the spontaneous growth and the *og* parameter is the organic growth. As can be seen, the *sng* and *og* values have a descending trend from the first scenario to the third scenario. This means the number of cells that were urbanized in undeveloped areas (*sng*) and the spread of development in urbanized areas (*og*) were decreasing from the first scenario to the third scenario. The *rt* parameter values, which represent the urban growth through roads and transportation networks, has not changed significantly between the three scenarios. These parameters prove that the growth rate in the unlimited condition was higher than the other scenarios and also in the compact scenario, which allowed the main body of the city to grow internally and externally, and limited the growth of suburb areas. This development was more than the last scenario, which only allowed rural suburb areas to be developed. The computed growth rate values in Table 6 confirm this fact and are minor in the third scenario than the other scenarios (Figure 6). Hence, due to the fact that at the moment, there is no particular policy in order to control the urban population and urban expansion and referring to increasing population in Tehran and in order to control the irregular urbanization and prevent from unfavorable consequences, the third scenario is considered as the most suitable scenario for Tehran province.

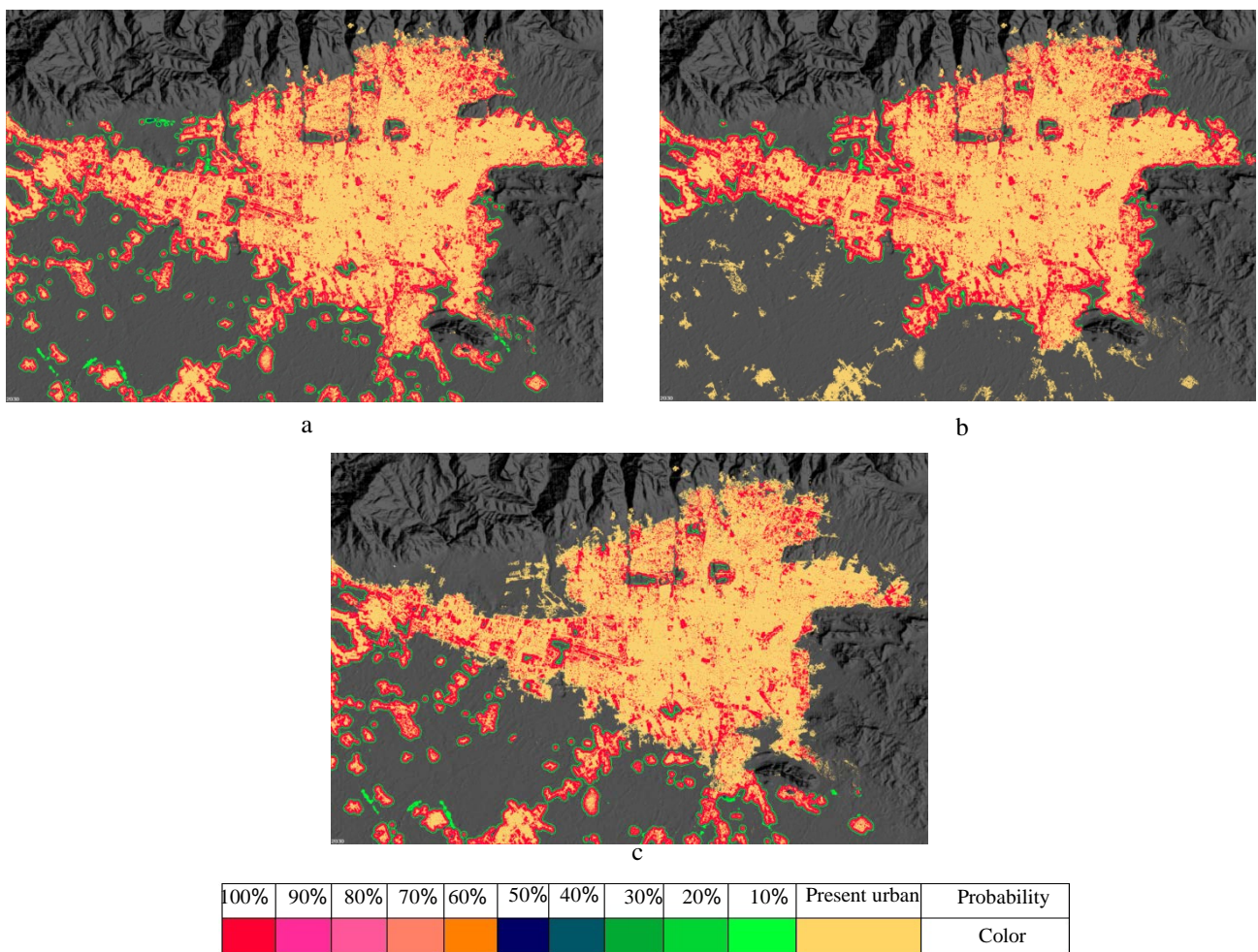

Figure 5 Urban growth probabilistic maps of year 2030 under three scenarios; (a) first scenario, (b) second scenario, (c) third scenario

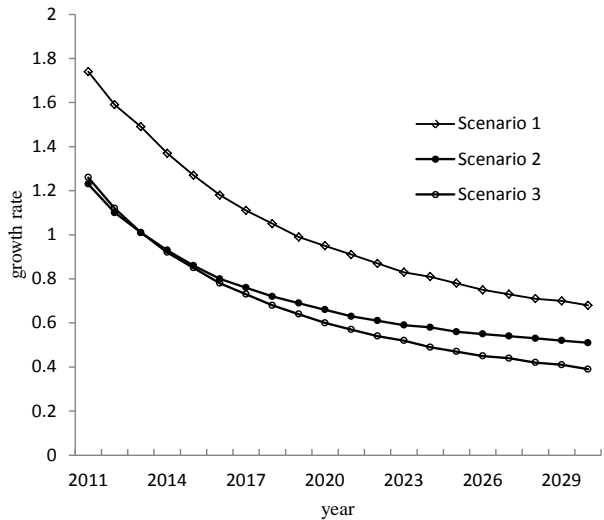

Figure 6 The growth rate of Tehran city within the prediction period under the three scenarios

## 4. CONCLUSION

As the population is increasing around the world from day to day, the requirement for the residential areas is increasing, and as a result, development in urbanized areas is accelerating. To prevent the unexpected consequences of irregular urbanization, the dynamic process of urban growth needs to be understood better. Understanding the dynamics of complex urban systems and evaluating the impact of urban growth on the environment involves procedures of modeling and simulation. Cellular Automata (CA) is a modeling procedure which is widely used by researchers of urban growth for predicting future urban extents.

In this research we used a CA based model called SLEUTH to simulate and predict the urban growth in Tehran province. SLEUTH is a dynamic, scale independent and future oriented model which can be applied to all regions under different conditions by modifying some initial conditions. It can also use different types of images simultaneously as input data and apply GIS environment and remote sensing tools to process them. Since there are many socioeconomic and environmental factors that affect the manner of the urbanization in cities and most of them are not available at suitable accuracies for planning and modeling purposes, there is no model that can be matched exactly with the reality. However, the three scenarios that were used in this research showed the usefulness of the SLEUTH model in order to compare the consequences of different conditions in a big city such as Tehran and provide valuable information about the manner of the urbanization in this city. Our research showed how Tehran city may develop in the future, under different growth and policy restrictions. This information can be used by city planners to better plan and prepare for future city-related issues and take action accordingly.

After preparing the input data, we calibrated and executed the model using three scenarios, which were different in the manner of restriction. In the first scenario, the only limitation was for highlands and deep slopes which were excluded from urbanization and initiated in the model, as an excluded layer. The second scenario prevented the rural suburb areas from development. In this scenario the urban growth was only occurring in the main body of the city and its interior developable regions. In the third scenario, the model was planned to encourage the urbanization in the rural areas. In this scenario, only the interior parts of the city had the possibility to be urbanized and the boundaries were excluded from urbanization. The results showed that organic growth ($og$) values decreased from year 2010 to year 2030 in the three scenarios. This means organic growth which appears from the existing urban pixels, had a decreasing trend over time. Additionally, small values of road influenced growth ($rt$), illustrate that transportation networks in Tehran had little effect on the manner of urban development. Moreover, small values of diffusion coefficient illustrated the minor scattering of urban regions in Tehran city.

On the other hand, both the spontaneous growth and the organic growth (*sng* and *og* respectively*)* values decreased sequentially from the first scenario to the third scenario. This means that both growth from the existing urban areas and growth in undeveloped areas in the third scenario are lower than the other two scenarios. Besides, overall, the growth rate was lower in the third scenario. Hence referring to conditions of Tehran as a big city, the last growth scenario was considered to be more suitable for this city in order to control the unexpected urbanization.